\documentclass[11pt]{article}
\usepackage[pdftex]{graphicx}
\usepackage{coling2018}
\usepackage{times}
\usepackage{url}
\usepackage{latexsym}

\usepackage{graphicx}
\usepackage{subcaption}
\usepackage{amsmath, amssymb}
\usepackage{booktabs}
\usepackage{multirow}
\usepackage{expex}

\title{Distance-Free Modeling of Multi-Predicate Interactions \\
in End-to-End Japanese Predicate-Argument Structure Analysis}

\author{
Yuichiroh Matsubayashi$^{\spadesuit\diamondsuit}$ \and Kentaro Inui$^{\spadesuit\diamondsuit}$ \\
{$^\spadesuit$Graduate School of Information Sciences, Tohoku University}\\
{$^\diamondsuit$RIKEN Center for Advanced Intelligence Project}\\
{\tt \{y-matsu, inui\}@ecei.tohoku.ac.jp}
}

\date{}

\begin{document}
\maketitle
\begin{abstract}
  Capturing interactions among multiple predicate-argument structures (PASs) is a crucial issue in the task of analyzing PAS in Japanese.
  In this paper, we propose new Japanese PAS analysis models that integrate the label prediction information of arguments in multiple PASs by extending the input and last layers of a standard deep bidirectional recurrent neural network (bi-RNN) model.
  In these models, using the mechanisms of pooling and attention, we aim to directly capture the potential interactions among multiple PASs, without being disturbed by the word order and distance.
  Our experiments show that the proposed models improve the prediction accuracy specifically for cases where the predicate and argument are in an indirect dependency relation and achieve a new state of the art in the overall $F_1$ on a standard benchmark corpus.
\end{abstract}

\section{Introduction}
\label{sec:intro}
\blfootnote{This work is licensed under a Creative Commons Attribution 4.0 International License. License
details: \url{http://creativecommons.org/licenses/by/4.0/}}

\begin{figure}[!h]
  \centering
  \includegraphics[width=0.55\linewidth]{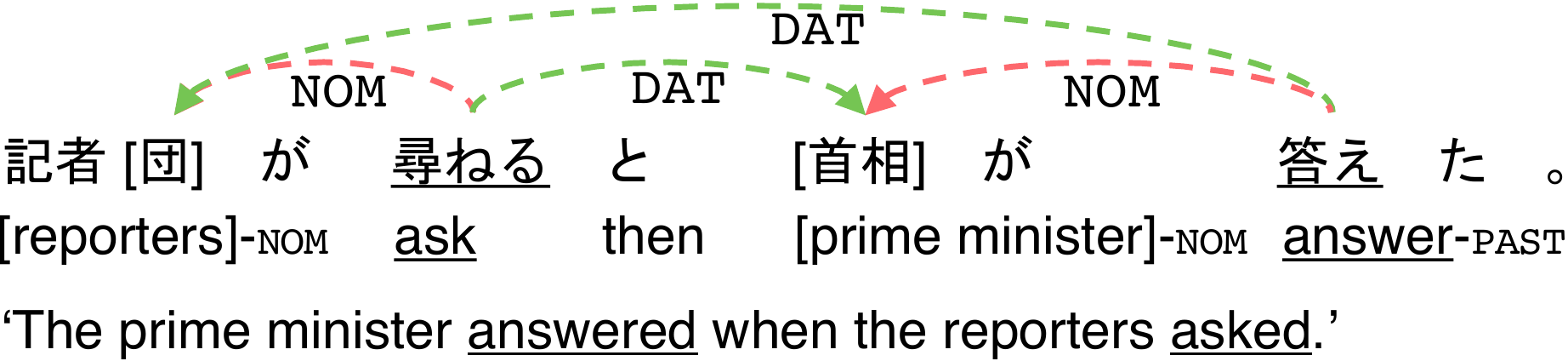}
  \caption{Example of PAS analysis. The dashed lines represent the predicate-argument relations. ``{\sf [reporters]}-{\sc NOM} {\sf ask then}'' constitutes a subordinate clause and ``{\sf [prime minister]}-{\sc NOM} {\sf answer}-{\sc past}'' constitutes a matrix clause.}
  \label{fig:example-sentence}
\end{figure}

A predicate-argument structure (PAS) is a structure that represents the relationships between a predicate and its arguments.
Identifying PASs in Japanese text is a long-standing challenge chiefly due to the abundance of omitted (elliptical) arguments.
In the example in Figure~\ref{fig:example-sentence}, the dative relation between {\it answer} and {\it reporters} is not explicitly indicated by the syntactic structure of the sentence.
We regard such arguments as elliptical and call those argument slots {\it Zero} cases.
$25\%$ of the obligatory arguments in Japanese newspaper articles are reported to be elliptical.\footnote{Statistics from the NAIST Text Corpus 1.5.~\cite{iida2017naist}}
The accuracy of identifying the fillers of such Zero cases remains only around 50\% in terms of $F_1$ even if the task is restricted to the identification of intra-sentential predicate-argument relations~\cite{Matsubayashi2017}.

One promising approach for addressing this problem is to model argument sharing across multiple predicates~\cite{Iida2015,Ouchi2015,ouchi2017}.
In Figure~\ref{fig:example-sentence}, for example, one can find very limited syntactic clues for predicting the long-distance dative relation between {\it answer} and {\it reporters}.
However, the relation must be easy to identify for human readers who know that {\it the person who asks a question is likely to be answered}; namely, the nominative argument of {\it ask} is likely to be shared with the dative argument of {\it answer}.
Capturing such inter-predicative dependencies has, therefore, been considered crucial of Japanese PAS analysis.

With this goal in mind, \newcite{Iida2015} constructed a {\it subject-shared predicate network} with an accurate recognizer of subject-sharing relations and deterministically propagated the predicted subjects to the other predicates in the graph.
However, this method is applied only to subject sharing, so it cannot take into account the relationships among multiple argument labels.

More recently, as an end-to-end model considering multi-predicate dependencies, \newcite{ouchi2017} used Grid RNN to incorporate intermediate representations of the prediction for one predicate generated by an RNN layer into the inputs of the RNN layer for another predicate.
However, in this model, since the information of multiple predicates also propagates through the RNNs, the integration of the prediction information is influenced by word order and distance, which is not necessarily important for aspects of syntactic and semantic relations.
Consequently, there might be information loss caused by the surface distances of words, as previous work had pointed out for RNN language models~\cite{Linzen2016}.

\begin{figure}[t]
  \begin{minipage}[b]{.48\linewidth}
    \centering
    \subcaptionbox{Base model \label{fig:base_model}}[\linewidth]{
    \includegraphics[width=\linewidth]{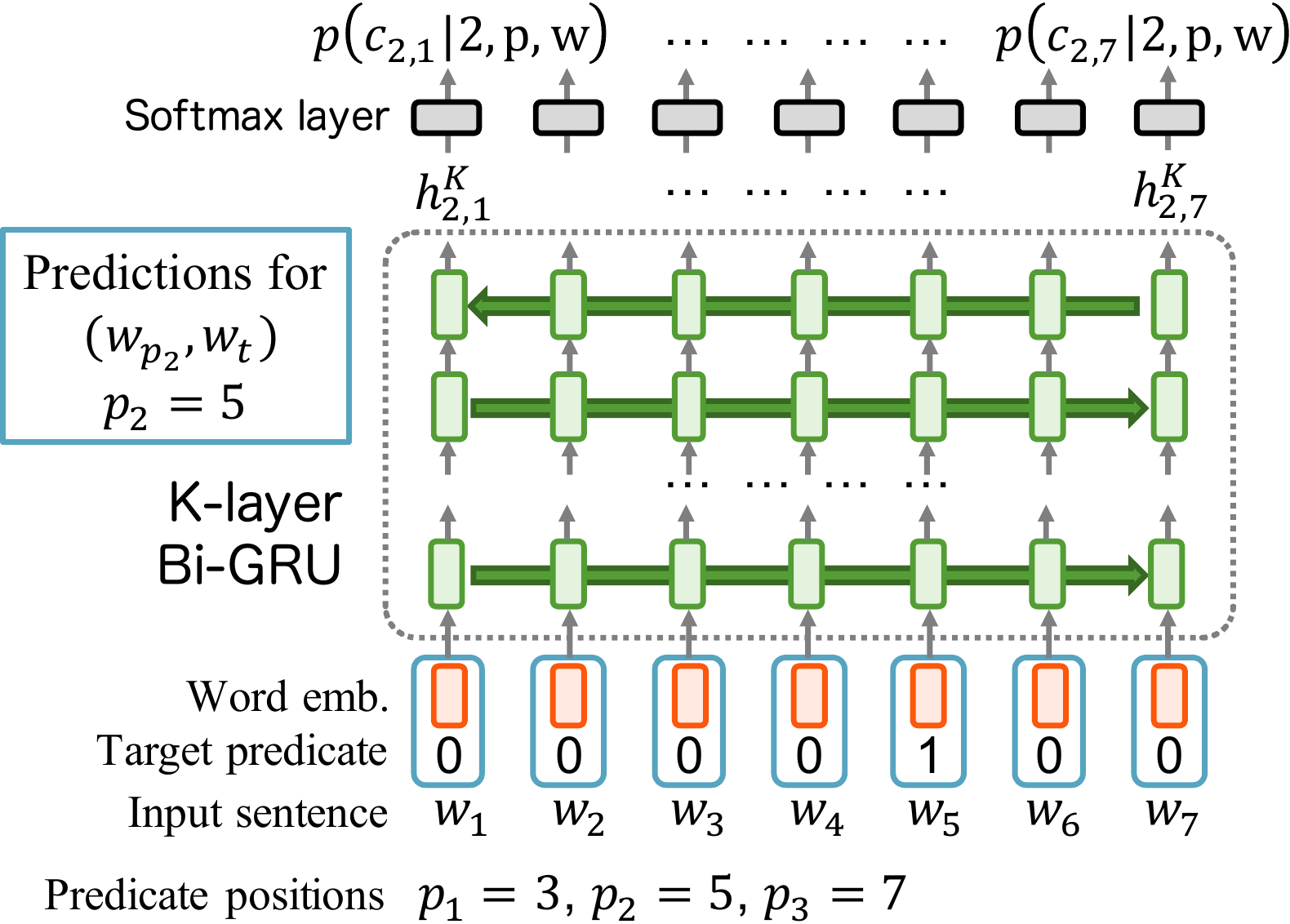}
    }%
  \end{minipage}
  \hspace{.01\linewidth}
  \begin{minipage}[b]{.5\linewidth}
    \centering
    \subcaptionbox{Proposed models \label{fig:our_model}}[\linewidth]{
    \includegraphics[width=\linewidth]{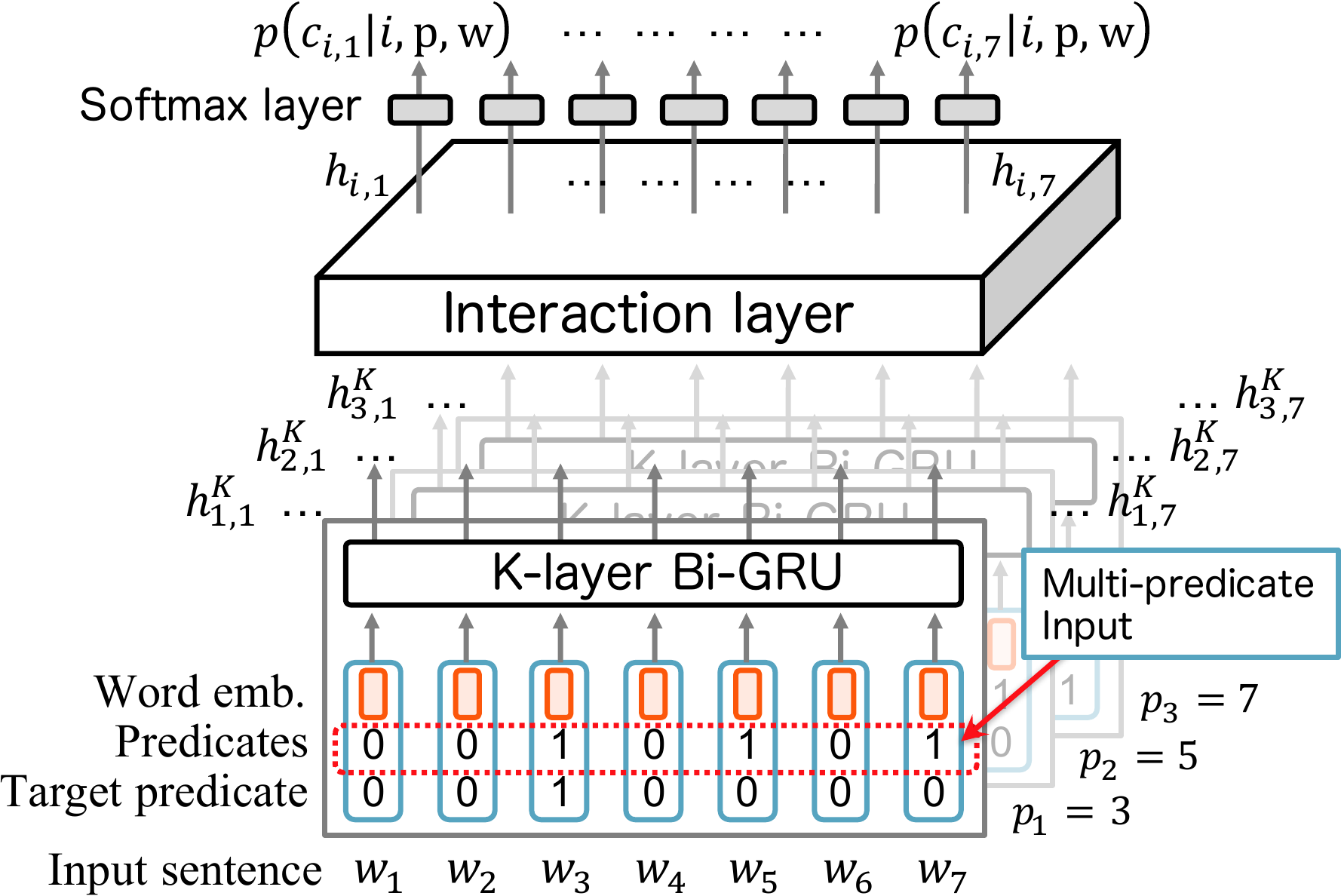}
    }
  \end{minipage}
  \caption{Network structures of the base and proposed models.}\label{tab:tab2}
\end{figure}

In this study, we propose new Japanese PAS analysis models that integrate the prediction information of arguments in multiple predicates.
We extend a standard end-to-end style deep bi-RNN model (Figure~\ref{fig:base_model}) and introduce components that consider the multiple predicate interactions into both the input and last layers (Figures~\ref{fig:our_model} and \ref{fig:our_extension}).
In contrast to Grid RNN, our extended models stack the extra layers using pooling and attention mechanisms on top of a deep bi-RNN so that they can directly associate the label prediction information for a target (predicate, word) pair with the predictions for words strongly related to the target pair.
Through experiments, we show that the proposed models improve argument prediction accuracy, especially for the {\it Zero} cases, and achieve a new state-of-the-art performance in the overall $F_1$ on a standard benchmark corpus.

\section{Task}
In this paper, we employ a task definition based on the NAIST Text Corpus (NTC)~\cite{iida2010annotation,iida2017naist}, a
commonly used benchmark corpus annotated with nominative ({\tt NOM}), accusative ({\tt ACC}), and dative ({\tt DAT}) arguments for predicates.
Given a tokenized sentence $\mathrm{w} = w_1, ... , w_n$ and its predicate positions $\mathrm{p}=p_1,...,p_q$,
our task is to identify at most one head of the filler tokens for each argument slot of each predicate.
In this study, we follow the setting of \newcite{Iida2015}, \newcite{ouchi2017}, and \newcite{Matsubayashi2017}, and focus only on analyzing arguments in a target sentence.
In addition, we exclude argument instances that are in the same {\it bunsetsu}, a base phrase unit in Japanese, as the target predicate, following \newcite{ouchi2017}, which we will compare with the results in experiments.

The semantic labels used in NTC may seem to be rather syntactic as they are named nominative, accusative, etc. However, this annotation task markedly differs from shallow syntactic parsing and is, in fact, more like a semantic role labeling (SRL) task including implicit argument prediction. First, the semantic labels in NTC generalize case alteration caused by voice alteration and thus represent semantic roles analogous to ARG0, ARG1, etc. in the PropBank-style annotation~\cite{palmer2005pba}. Second, in the corpus, when an argument is omitted (i.e., zero-anaphora), the antecedent is identified with an appropriate semantic role, which is a prominent problem in Japanese semantic analysis and is the primary target of this study.

\section{Base Model}

Our proposed models extend end-to-end style SRL systems using deep bi-RNN~\cite{Zhou2015,He2017,ouchi2017} to combine mechanisms that consider multiple predicate interactions.
Figure~\ref{fig:base_model} shows the network of our base model.
Formally, given a word sequence $\mathrm{w} = w_1, ... , w_n$ and a target predicate position $p_i$ in $\mathrm{p}$,
the model outputs a label probability for each word position: $p(c_{i,1}|i, \mathrm{p}, \mathrm{w}), ... , p(c_{i,n}|i, \mathrm{p}, \mathrm{w})$.
Here, $c_{i,t} \in \{\mathtt{NOM}, \mathtt{ACC}, \mathtt{DAT}, \mathtt{NONE}\}$ represents the argument label of the word $w_t$ for the target predicate $w_{p_i}$.

The input layer creates a vector $h^{0}_{i,t} \in \mathbb{R}^{d_w+1}$ for each pair of a predicate $w_{p_i}$ and a word $w_t$ by concatenating a word embedding $e(w_t) \in \mathbb{R}^{d_w}$ and a binary value representing the target predicate position in a method similar to that of \newcite{He2017}.
The obtained vectors are then input into the deep bi-RNN, where the directions of the layers alternate~\cite{Zhou2015}:
\begin{align}
  h^1_{i,t} &= r^1(h^{0}_{i,t}, h^{1}_{i,t-1}), \quad
  h^k_{i,t} =
  \begin{cases}
    h^{k-1}_{i,t} + r^k(h^{k-1}_{i,t}, h^{k}_{i,t-1}) & (k \mbox{ is odd})\\
    h^{k-1}_{i,t} + r^k(h^{k-1}_{i,t}, h^{k}_{i,t+1}) & (k \mbox{ is even})
  \end{cases} \label{eqn:base-model2}
  (k \geq 2).
\end{align}
Here, $h^{k}_{i,t} \in \mathbb{R}^{d_r}$ is the output of the $k$-th RNN layer for a pair ($w_{p_i}$, $w_t$), and $r^k$ is a function representing the $k$-th RNN layer.
We employ gated recurrent units (GRUs)~\cite{Cho2014} for the RNNs.
In addition, we use the residual connections~\cite{He2016} following \newcite{ouchi2017}.
Then, a four-dimensional vector representing a probability $p(c_{i,t}|i, \mathrm{p}, \mathrm{w})$ is obtained by applying a softmax layer to each output of the last RNN layer $h^K_{i,t}$.
For each argument label $c$ of each predicate, we eventually select a word with the maximum probability that exceeds an output threshold $\theta_c$.

\section{Proposed Models}
\label{sec:model}

\begin{figure}[t]
  \centering
  \includegraphics[width=\linewidth]{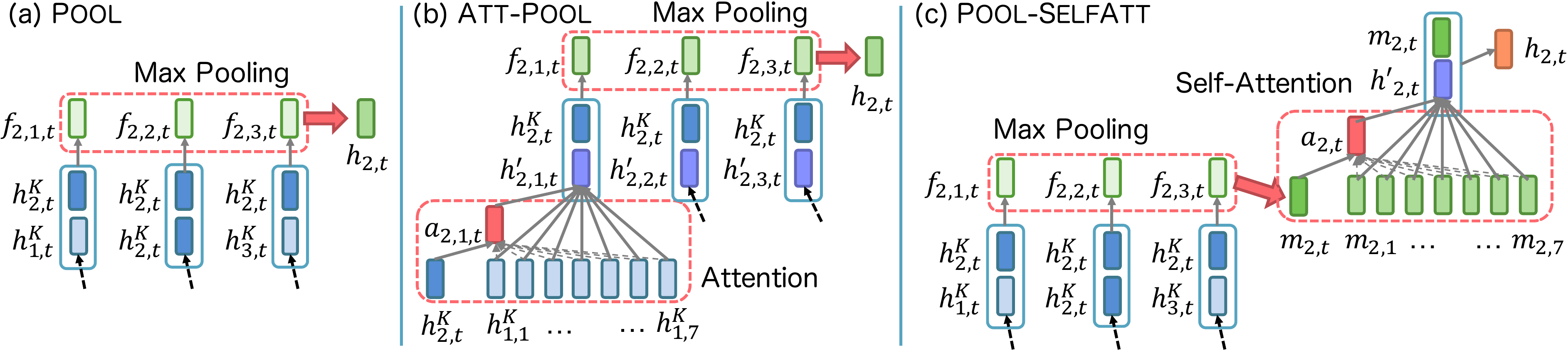}
  \caption{Three variants of interaction layers.}
  \label{fig:our_extension}
\end{figure}

Our base model independently predicts the arguments of each predicate.
In order to capture dependencies between the arguments of multiple predicates, we apply two extensions to our base model: a {\it multi-predicate input layer} and three variants of {\it interaction layers} on top of the deep bi-RNNs.
Figures~\ref{fig:our_model}~and~\ref{fig:our_extension} show the network structures of the extended models.

In contrast to the Grid RNN model of \newcite{ouchi2017}, where the information of multiple predicates propagates through the RNNs, our interaction layers use pooling and attention mechanisms to directly associate the label prediction information for a target (predicate, word) pair with that for words strongly related to the target pair, without being disturbed by word order and distance.

\subsection{Interaction Layers}

\paragraph{Pooling ({\sc Pool})}

Argument sharing across multiple predicates can be captured with both syntactic and semantic clues.
At the syntactic level, we want to capture tendencies that, for example, the subject of the predicate of a matrix clause is likely to fill argument slots of other predicates in the same sentence.
At the semantic level, we want to model semantic dependencies between neighboring events such as {\it the person who asks a question is likely to be answered}, as in Figure~\ref{fig:example-sentence}.
Our proposal is to capture both types of clues by incorporating a max pooling layer on top of the base model.

Specifically, as illustrated in Figure~\ref{fig:our_extension}a, for each word $w_t$,
we integrate the intermediate representation of label prediction for each predicate $h^K_{i,t}$ by applying max pooling to the vectors
that represent pairs of prediction information for two predicates $h^K_{i,t}$ and $h^K_{j,t}$
(including the case $i=j$):
\begin{align}
  h_{i,t}   &= \mathrm{maxpool}_{j}(f_{i,j,t}), \quad \mbox{where} \quad
  f_{i,j,t} = \mathrm{ReLU}(W_f[h^K_{i,t},h^K_{j,t}] + b_f).
\end{align}
In this equation, $\mathop{\mathrm{maxpool}}_{j}(f_{i,j,t})$ is an operation to extract the maximum value of each dimension in $\{f_{i,1,t},...,f_{i,q,t}\}$.
The newly obtained vector $h_{i,t}$ for $w_{p_i}$ and $w_t$ is input into the softmax layer in the same manner as in the base model.

\paragraph{Attention-then-Pooling ({\sc Att-Pool})}

Besides the argument sharing across multiple predicates,
we would also like to capture dependencies between different arguments of a single predicate (and potentially, arguments of multiple predicates).
For example, syntactically, two distinct argument slots of a single predicate are unlikely to share the same filler.
Semantically, the subject of a predicate {\it take} is likely to be a person when its object is {\it a bread}, but is likely to be a company if the object is {\it a new employee}.

To capture such dependencies, we integrate the intermediate label prediction $h^K_{j,t'}$ of $w_{t'}$ for an arbitrary predicate $w_{p_j}$ (including the case $i=j$) into the prediction of $w_{t}$ for a target predicate $w_{p_i}$.
In the integration, we aim to weigh the prediction information for $(w_{p_j}, w_{t'})$ based on its relatedness to the target pair $(w_{p_i}, w_{t})$ using the attention mechanism~\cite{Bahdanau2014}.
As in Figure~\ref{fig:our_extension}b, we calculate a weight $a_{i,j,t}(t') \in \mathbb{R}$ for each of $h^K_{j,1}, ..., h^K_{j,n}$ on the basis of the prediction $h^K_{i,t}$ for the target pair and we obtain a weighted sum of $h^K_{j,t'}$ as a summary of the argument information of $w_{p_j}$, which is expected to be useful for the label prediction of $(w_{p_i}, w_{t})$:
\begin{align}
  h'_{i,j,t}    &= \textstyle{\sum}_{t'}{a_{i,j,t}(t') \cdot h^K_{j,t'}}, \quad \mbox{where} \quad
  a_{i,j,t}(t') = \frac{\exp(W_a g_{i,j,t,t'} + b_a)}{\sum_{t''}\exp(W_a g_{i,j,t,t''} + b_a)}, \\
  g_{i,j,t,t'}  &= \tanh(W_g[h^K_{i,t},h^K_{j,t'}]+b_g).
\end{align}
The obtained $h'_{i,j,t}$ are concatenated with the prediction for the target pair $h^K_{i,t}$ and linearly transformed with the ReLU activation.
Max pooling is then applied to these vectors to combine the predictions for multiple predicates.
\begin{align}
  h_{i,t} &= \mathrm{maxpool}_{j}(f_{i,j,t}), \quad \mbox{where} \quad
  f_{i,j,t} = \mathrm{ReLU}(W_f[h^K_{i,t},h'_{i,j,t}] + b_f)
\end{align}

\paragraph{Pooling-then-Self-Attention ({\sc Pool-SelfAtt})}
The {\sc Att-Pool} model involves a high computational cost because it must compute $nq^2$ different attentions regarding the number of words $n$ and the number of predicates $q$ in a sentence.
Therefore, as illustrated in Figure~\ref{fig:our_extension}c, in this model, we first apply the max pooling that we applied in the {\sc Pool} model to reduce the sequences for which attentions must be computed by integrating the label predictions of $w_t$ for all the other predicates in advance.
\begin{align}
  m_{i,t}     &= \mathrm{maxpool}_{j}(f_{i,j,t}), \quad \mbox{where} \quad
  f_{i,j,t}   = \mathrm{ReLU}(W_f[h^K_{i,t},h^K_{j,t}] + b_f)
\end{align}
Then, we combine the information in the obtained sequence $m_{i,1}, ... ,m_{i,n}$ in a similar manner as in the {\sc Att-Pool} model using the attention mechanism, but this time, with self-attention, that is, computing the weights of the elements in the sequence based on the relatedness to the element inside the sequence.
\begin{align}
  h_{i,t}     &= \mathrm{ReLU}(W_h[m_{i,t},h'_{i,t}] + b_h) \\
  h'_{i,t}    &= \textstyle{\sum_{t'}}{a_{i,t}(t') \cdot m_{i,t'}}, \quad \mbox{where} \quad
  a_{i,t}(t') = \frac{\exp(W_a g_{i,t,t'} + b_a)}{\sum_{t''}\exp(W_a g_{i,t,t''} + b_a)} \\
  g_{i,t,t'}  &= \tanh(W_g[m_{i,t},m_{i,t'}]+b_g)
\end{align}
Consequently, the number of attentions that must be computed is reduced to $nq$.

\paragraph{Self-Attention ({\sc SelfAtt})}
To conduct ablation tests to assess the impact of the proposed extensions, we also implemented a model only with self-attention.
This model explicitly considers the relationships between arguments of a single predicate, but not arguments across multiple predicates.
\begin{align}
  h_{i,t}     &= \mathrm{ReLU}(W_h[h^K_{i,t},h'_{i,t}] + b_h) \\
  h'_{i,t}    &= \textstyle{\sum}_{t'}{a_{i,t}(t') \cdot h^K_{i,t'}}, \quad \mbox{where} \quad
  a_{i,t}(t') = \frac{\exp(W_a g_{i,t,t'} + b_a)}{\sum_{t''}\exp(W_a g_{i,t,t''} + b_a)} \\
  g_{i,t,t'}  &= \tanh(W_g[h^K_{i,t},h^K_{i,t'}]+b_g)
\end{align}

\subsection{Multi-Predicate Input Layer ({\sc MP})}
In addition, we add a simple but effective extension to the input layer.
As \newcite{He2016} reported, the information of the target predicate $w_{p_i}$ propagates to the intermediate prediction $h^K_{i,t}$ of the candidate argument $w_t$ through the deep bi-RNN by just adding a binary value representing the predicate position.
Inspired by this finding, as shown in Figure~\ref{fig:our_model},
in the input layer, we add another binary value that represents all the predicate positions to $h^{0}_{i,t}$, aiming to propagate multiple predicate information.

\section{Experiments}
We evaluated the impacts of our extensions and compared their performances to those of previous studies.
Our main hypothesis is that the pooling and attention mechanisms are both useful for capturing different types of argument interactions as we explained in Section~\ref{sec:model} and do work complementarily of each other to improve the prediction accuracy, especially for arguments in a long-distance dependency.

\subsection{Settings}
\subsection{Dataset and Implementation Details}
The experiments were performed on NTC 1.5.
We divided the corpus into the commonly used divisions of training, development, and test sets~\cite{taira2008japanese}, each of which includes 24,283, 4,833, and 9,284 sentences, respectively.
NTC represents each argument of a predicate by indicating a coreference cluster in a text.
For each given predicate-argument slot, we count a system's output as correct if the output token is included in the coreference cluster corresponding to the slot fillers.
The evaluation is performed on the basis of the precision, recall, and $F_1$ score.

The hyperparameters were selected to obtain a maximum $F_1$ on the development set.
The details of the hyperparameter selection and preprocessing are described in the supplemental material.
In the following experiments, we train each model 10 times with the same training data and hyperparameters and then show the average scores.

\subsection{Grid RNN Baseline ({\sc Grid})}
\label{sec:grid-rnn}
In order to strictly compare the impact of our extensions to the method used for integrating multiple pieces of predicate information in the state-of-the-art end-to-end model, in addition to our base model, we replicated the method of \newcite{ouchi2017} by modifying
Equations~(\ref{eqn:base-model2})
 of our base model as follows:
\begin{align}
  h^1_{i,t} &= r^1([h^{0}_{i,t},h^{1}_{i-1,t}], h^{1}_{i,t-1}), \quad
  h^k_{i,t} = h^{k-1}_{i,t} +
\begin{cases}
  r^k([h^{k-1}_{i,t},h^{k}_{i-1,t}], h^{k}_{i,t-1}) & (k \mbox{ is odd})\\
  r^k([h^{k-1}_{i,t},h^{k}_{i+1,t}], h^{k}_{i,t+1}) & (k \mbox{ is even})
\end{cases}
(k \geq 2),
\end{align}
if $1 \leq i \leq q$; otherwise, $h^{k}_{i,t} = {\bf 0}$. The performance of this replicated model may not be strictly the same as that reported in \newcite{ouchi2017} due to discrepancies in the embeddings of inputs, hyperparameters (a training batch size, a hidden unit size, etc.), and training strategy (an optimizing algorithm, a regularization method, an early stopping method, etc.).
The predicate positions $\mathrm{p}=p_1,...,p_q$ are arranged in ascending order.

 \begin{table*}[!t]
   \centering
   \scriptsize
   \begin{tabular}{l|l||rr||r|r|r|r|r|r|r|r}
     \toprule
     &  & \multicolumn{4}{c|}{All} & \multicolumn{6}{c}{$F_1$ at different dependency distances} \\
     \cmidrule(){2-12}
     &  Model & $F_1$ (\%)& $SD$ & Prec. & Rec. & {\it Dep} & {\it Zero} & 2 & 3 & 4 &  $\geq$ 5 \\
     \midrule
     & {\sc Base } ($d_r=32$, $K=8$)          &  81.22 & $\pm$0.19        & 84.30 & 78.37 & 88.39 & 49.12 & 55.73 & 47.1 & 39  & 29 \\
     Baseline & {\sc Grid } ($d_r=32$, $K=8$) &  81.06 & $\pm$0.31        & 84.33 & 78.04 & 88.17 & 48.73 & 55.26 & 47.5 & 39  & 28 \\
     Models & {\sc Base       }               &  83.39 & $\pm$0.13        & 85.85 & 81.07 & 89.90 & 54.37 & 61.09 & 53.8 & 44  & 31 \\
     & {\sc Grid       }                      &  82.94 & $\pm$0.17        & 85.38 & 80.63 & 89.51 & 53.57 & 60.28 & 52.4 & 44  & 32 \\
     \midrule
     & {\sc SelfAtt    }                      &  83.56 & $\pm$0.22        & 85.91 & 81.34 & 90.06 & 54.84 & 61.36 & 54.3 & {\bf 45}  & 32 \\
     & {\sc Pool       }                      &  83.56 & $\pm$0.16        & 86.05 & 81.21 & 90.00 & 54.81 & 61.54 & 54.3 & {\bf 45}  & 31 \\
     Proposed & {\sc Att-Pool}                &  83.48 & $\pm$0.24        & 85.97 & 81.12 & 89.98 & 54.57 & 61.19 & 54.0 & 44  & 32 \\
     Models & {\sc Pool-SelfAtt}              &  83.76 & $\pm$0.17        & 86.11 & 81.54 & 90.17 & 55.19 & 62.10 & 54.0 & {\bf 45}  & 32 \\
     & {\sc MP         }                      &  83.67 & $\pm$0.22        & 86.08 & 81.39 & 90.10 & 54.80 & 61.67 & 53.8 & 44  & 32 \\
     & {\sc MP-SelfAtt }                      &  83.79 & $\pm$0.22        & 86.11 & {\bf 81.60} & 90.22 & 55.26 & 61.88 & 54.3 & {\bf 45}  & {\bf 33} \\
     & {\sc MP-Pool-SelfAtt}                  &  {\bf 83.94} & $\pm$0.12 & {\bf 86.58} & 81.46 & {\bf 90.26} & {\bf 55.55} & {\bf 62.44} & {\bf 54.7} & {\bf 45}  & 32 \\
     \midrule
     Previous & \newcite{ouchi2017}        & 81.42 &              & & & 88.17 & 47.12 & &&&\\
     SOTAs & M\&I 2017 & 83.50 & $\pm$0.17  & 86.00 & 81.15 & 89.89 & 51.79 & 60.17 & 49.4 & 38  & 23 \\
     \midrule
     \midrule
     Ensemble & {\sc MP-Pool-SelfAtt} (10 models)  & {\bf 85.34} & & {\bf 87.90} & {\bf 82.93} & {\bf 91.26} & {\bf 58.07} & {\bf 64.89} & {\bf 57.5} & {\bf 47} & {\bf 33} \\
     Models & M\&I 2017 (5 models)   & 84.07 &     & 86.09 & 82.15 & 90.24 & 53.66 & 61.94 & 51.8 & 40  & 24 \\
     \bottomrule
   \end{tabular}
   \caption{$F_1$ scores on the NTC 1.5 test set. {\it Dep} and {\it Zero} denote instances where the dependency distance between the predicate and argument is one and more than one, respectively. M\&I 2017 is the model of \newcite{Matsubayashi2017}.
   }
   \label{tbl:model-comparison}
 \end{table*}

\begin{table*}[!t]
\centering
\scriptsize
\begin{tabular}{l|rr|r|r|r|r|r|r|r}
\toprule
\multicolumn{1}{r|}{Model B}&& 	& 	{\sc Base } 	& 	{\sc Att- } 	& 	{\sc SelfAtt } 	& 	{\sc Pool } 	& 	{\sc MP } 	& 	{\sc Pool- } 	& 	{\sc MP- } 	\\
Model A 	& $F_1$ (\%) & $SD$ & 	{\sc  } 	    & 	{\sc Pool } 	& 	{\sc  }  	& 	{\sc } 	& 	{\sc } 	& 	{\sc SelfAtt } 	& 	{\sc SelfAtt } 	\\
\midrule
{\sc Base } 	         & 83.39 & $\pm$0.13 \\
{\sc Att-Pool } 	     & 83.48 & $\pm$0.24 & 0.18 \\
{\sc SelfAtt } 	       & 83.56 & $\pm$0.22 & {\bf 0.03}   & 0.22 	\\
{\sc Pool } 	         & 83.56 & $\pm$0.16 & {\bf 0.014}  & 0.21   	& 0.53 \\
{\sc MP } 	           & 83.67 & $\pm$0.22 & {\bf 0.003}  & {\bf 0.048}  & 0.16 	  & 0.12 \\
{\sc Pool-SelfAtt } 	 & 83.76 & $\pm$0.17 & {\bf 4.3E-5} & {\bf 0.004}  & {\bf 0.023}  & {\bf 0.0084} & 0.16 \\
{\sc MP-SelfAtt }      & 83.79 & $\pm$0.22 & {\bf 1.0E-4} & {\bf 0.0046} & {\bf 0.021}  & {\bf 0.0096} & 0.13 	 & 0.39 	\\
{\sc MP-Pool-SelfAtt } & 83.94 & $\pm$0.12 & {\bf 5.4E-6} & {\bf 5.4E-6} & {\bf 2.2E-4} & {\bf 2.7E-5} & {\bf 0.0013} & {\bf 0.013} & {\bf 0.046} \\
\bottomrule
\end{tabular}
\caption{{\it p-values} in one-sided permutation test using 10 overall $F_1$ scores for each model.
The bold values indicate that an average $F_1$ score of model A outperforms that of model B at the $5$\% significance level.}
\label{tbl:significance-test}
\end{table*}

\subsection{Results}
\subsubsection*{Impact of Extensions}

The first two sets of rows in Table~\ref{tbl:model-comparison} compare the impact of each component of our extension.
The effects of incorporating the interaction layer can be seen in the comparisons of the {\sc Base} model with the {\sc SelfAtt}, {\sc Pool}, {\sc Att-Pool}, and {\sc Pool-SelfAtt} models.
Among the four proposed extensions, {\sc Pool-SelfAtt}, an integration of {\sc Pool} and {\sc SelfAtt}, achieved the best performance ($83.76$  in $F_1$), gaining $0.37$ points in overall $F_1$ from {\sc Base}.
Also, the significance tests in Table~\ref{tbl:significance-test} show that the {\sc Pool} and {\sc SelfAtt} models significantly outperform the BASE model, and the {\sc Pool-SelfAtt} model makes a further significant gain from the {\sc Pool} and {\sc SelfAtt} models.
This indicates that {\sc Pool} and {\sc SelfAtt} work complementarily with each other, and combining them makes a further improvement from each individual extension.
Recall that {\sc SelfAtt} is designed to capture long-distance dependencies over a single predicate-argument structure, whereas {\sc Pool} is expected to capture argument sharing across multiple predicates.
These results provide empirical support to the hypotheses behind our design of the interaction layer.

The {\sc MP} model, where the input layer is extended to represent the positions of all the predicates in a sentence, significantly outperforms the {\sc Base} model by $0.28$ points in overall $F_1$.
This result suggests the importance of position information regarding the neighboring predicates in identifying the arguments of a given predicate.
Furthermore, the {\sc MP-Pool-SelfAtt} model, which is a combination of {\sc MP} and {\sc Pool-SelfAtt}, resulted in a further 0.27-point improvement and consequently achieved the best overall $F_1$ of 83.94 as a single model.

Following \newcite{Matsubayashi2017}, we also assess $F_1$ values at different dependency distances. The results are shown in the right half of Table~\ref{tbl:model-comparison}.
From the table, we can see that {\sc MP-Pool-SelfAtt} improves $F_1$ from {\sc BASE} by $0.9$--$1.4$ points consistently across all the distance categories other than {\it Dep}.

\begin{table*}[t]
  \scriptsize
  \centering
  \begin{tabular}{cc}%
    \begin{tabular}{l|l|rrrr|rrrr}
      \toprule
      & & \multicolumn{4}{c|}{Dep} & \multicolumn{4}{c}{Zero}\\
      Model & ALL & ALL & NOM & ACC & DAT & ALL & NOM & ACC & DAT \\
      \midrule
      {\sc MP-Pool-SelfAtt} & {\bf 83.94} & {\bf 90.26} & 90.88 & 94.99 & {\bf 67.57} & {\bf 55.55} & {\bf 57.99} & {\bf 48.9} & {\bf 23} \\
      \newcite{Ouchi2015}        & 79.23 & 86.07 & 88.13 & 92.74 & 38.39 & 44.09 & 48.11 & 24.4 & 4.8 \\
      \newcite{ouchi2017}        & 81.42 & 88.17 & 88.75 & 93.68 & 64.38 & 47.12 & 50.65 & 32.4 & 7.5 \\
      M\&I 2017 & 83.50 & 89.89 & {\bf 91.19} & {\bf 95.18} & 61.90 & 51.79 & 54.69 & 41.8 & 17  \\
      \midrule
      \midrule
      {\sc MP-Pool-SelfAtt} (ens.)               & {\bf 85.34} & {\bf 91.26} & {\bf 91.84} & {\bf 95.57} & {\bf 70.8 } & {\bf 58.07} & {\bf 60.21} & {\bf 52.5} & {\bf 26}\\
      M\&I 2017 (ens. of 5) & 84.07 & 90.24 & 91.59 & 95.29 & 62.61 & 53.66 & 56.47 & 44.7 & 16 \\
      \bottomrule
    \end{tabular}
    &
    \begin{tabular}{l|r}
      \toprule
      \multicolumn{2}{c}{Modified NTC 1.5} \\
      \multicolumn{2}{c}{\cite{Iida2016}}  \\
      \midrule
      \multirow{2}{*}{Model}      & \multicolumn{1}{c}{Zero}  \\
      & NOM \\
      \midrule
      \newcite{Ouchi2015} & {\bf 57.3} \\
      \newcite{Iida2015}  & 41.1 \\
      \newcite{Iida2016}  & 52.5 \\
      \bottomrule
      \multicolumn{2}{l}{(Note) Results on a dataset}  \\
      \multicolumn{2}{l}{different from our experiments}  \\
    \end{tabular}
  \end{tabular}
  \caption{$F_1$ scores of each argument label on the NTC 1.5 test set.}
  \label{tbl:result-each-case}
\end{table*}

\subsubsection*{Comparison to Related Work}
The third set of rows in Table~\ref{tbl:model-comparison} shows the reported performance of related studies.
Grid RNN of \newcite{ouchi2017} is a state-of-the-art end-to-end model, designed to capture interactions among multiple predicate-argument relations.
A comparison between their model and the proposed models was somewhat tricky because our replication of Grid RNN did not reproduce the reported performance on the same dataset (see the row of {\sc Grid} in Table~\ref{tbl:model-comparison}).
Unlike the results reported in \newcite{ouchi2017}, the {\sc Grid} model in our experiment did not clearly outperform the model without the grid architecture, i.e., the {Base} model.
We first suspected that this might have resulted from the difference in dimensionality $d_r$ of RNN hidden states: $d_r=32$ in \newcite{ouchi2017}, whereas $d_r=256$ in our experiments.
Specifically, we speculated that the base model with a low dimensionality left a larger margin for improvement and incorporating the Grid architecture derived positive effects.
We thus trained our {\sc Grid} model with \newcite{ouchi2017}'s settings ($d_r=32$ and $K=8$) and the best performing hyperparameters; however, we were not able to reproduce the reported gain from Grid RNN (see the row of ``{\sc Grid} ($d_r=32$, $K=8$)'' in Table~\ref{tbl:model-comparison}).\footnote{We discussed this negative result, including the implementation details, with one of the authors of \newcite{ouchi2017}. However, we could not find a plausible reason for the results.}
This might be an indication of the difficulty in capturing multi-predicate interactions by threading deep bi-RNNs with RNNs, as we discussed in Section~\ref{sec:intro}.

Another previous state-of-the-art model was proposed by \newcite{Matsubayashi2017} ({\sc M\&I~2017}).
This model extends a feedforward NN with dependency path embeddings and other new features to capture long-distance dependencies in a single PAS.
The row ``{\sc M\&I 2017}'' in Table~\ref{tbl:model-comparison} shows the reported performance of their model.\footnote{For the purpose of a strict comparison with \newcite{ouchi2017}, we re-evaluate the model of \newcite{Matsubayashi2017} by excluding the instances for which the argument is in the same {\it bunsetsu} phrase as the predicate; this is the same setting as that in \newcite{ouchi2017}. We have reported the new results in Tables~\ref{tbl:model-comparison} and \ref{tbl:result-each-case}.}
The performance of {\sc M\&I 2017} is comparable with the performance of our {\sc SelfAtt} model.
This result provides another piece of empirical evidence that the self-attention mechanism has a comparably positive effect in incorporating dependency path information for capturing long-distance dependencies in a single PAS.

Overall, the proposed methods of using the pooling and attention mechanisms for capturing interactions across predicates and arguments gained considerable improvement and achieved state-of-the-art accuracy, significantly outperforming the previous state-of-the-art models.
The last set of rows in Table~\ref{tbl:model-comparison} shows the results of the ensemble models.
A model that predicts arguments with the average score of the 10 {\sc MP-Pool-Att} models further improves the overall $F_1$ by 1.4 points from that of a single model, achieving state-of-the-art accuracy for NTC 1.5.

Table~\ref{tbl:result-each-case} shows the $F_1$ score for each case label.
In a comparison of the single models, although our {\sc MP-Pool-Att} model slightly degrades the scores of {\tt NOM} and {\tt ACC} on the {\it Dep} cases compared to the state-of-the-art model ({\sc M\&I 2017}), it greatly improves the scores for {\tt DAT} and the {\it Zero} cases.
Regarding the ensemble models, {\sc MP-Pool-Att} improves the scores for all cases.

\newcite{Iida2015} and \newcite{Iida2016} report Japanese subject anaphora resolution systems, designed to predict only {\it Zero} NOM arguments.
It is not straightforward to directly compare their results with ours due to the differences in the experimental settings.
However, our best performing model outperforms the model of \newcite{Ouchi2015}, which is then reported to outperform both \newcite{Iida2015} and \newcite{Iida2016} in their experimental settings.

\begin{figure}[!t]
  \centering
  \includegraphics[width=\linewidth]{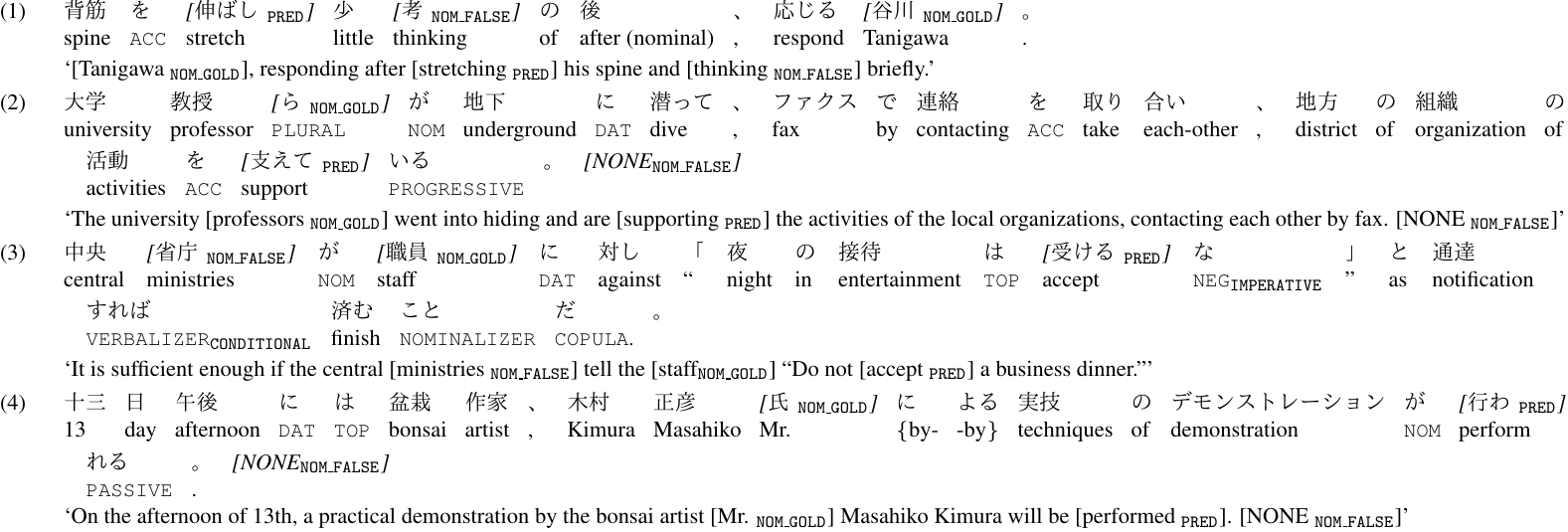}\\
  \caption{Examples of prediction errors. In Example (1), only {\sc SelfAtt} failed to predict the answer. In Example (2), only {\sc MP-Pool-SelfAtt} correctly predicted the answer. In Examples (3) and (4), none of the systems predict the answers correctly.}
  \label{fig:analysis}
\end{figure}

\begin{figure}[t]
  \centering
  \includegraphics[width=0.69\linewidth]{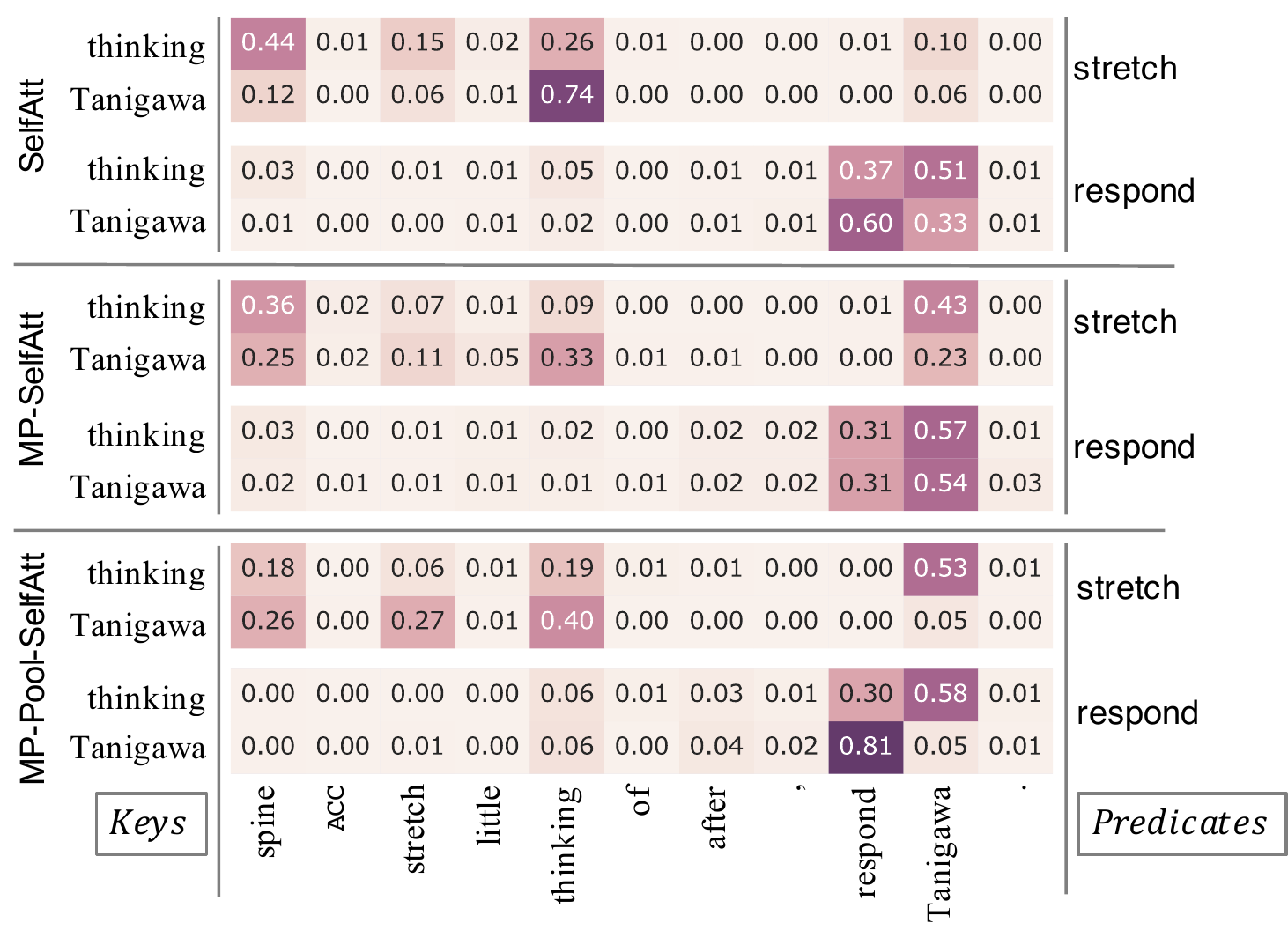}\\
  \caption{Attention weights of proposed models for Example~(1).}
  \label{fig:attention-ex1}
\end{figure}

\subsection{Detailed Analysis}
\begin{figure*}[!t]
  \centering
  \includegraphics[width=\linewidth]{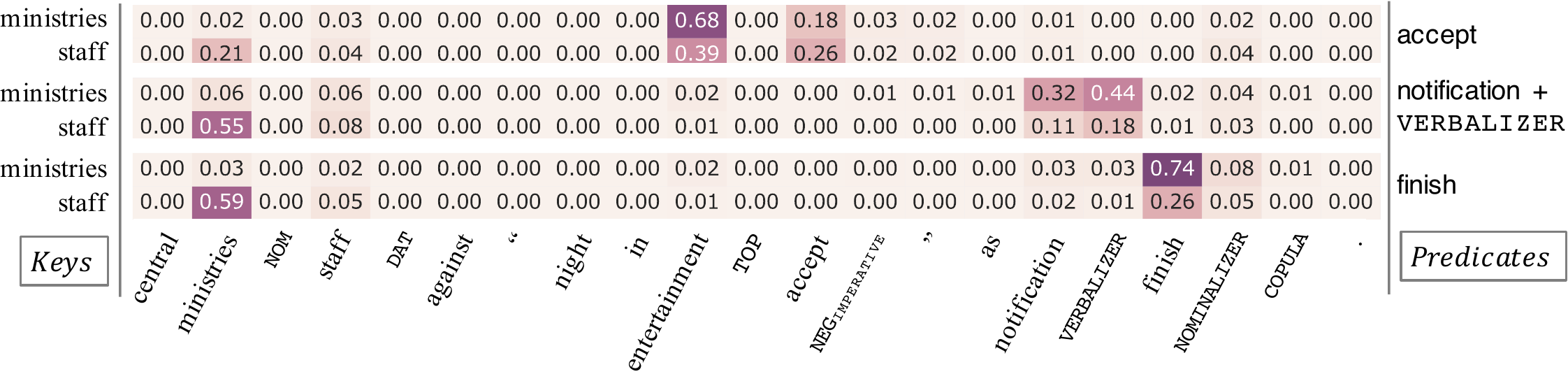}\\
  \caption{Attention weights of {\sc MP-Pool-SelfAtt} for Example~(3).}
  \label{fig:attention-ex3}
\end{figure*}

\begin{figure*}[!t]
  \centering
	\includegraphics[bb=0 20 682 155,width=\linewidth]{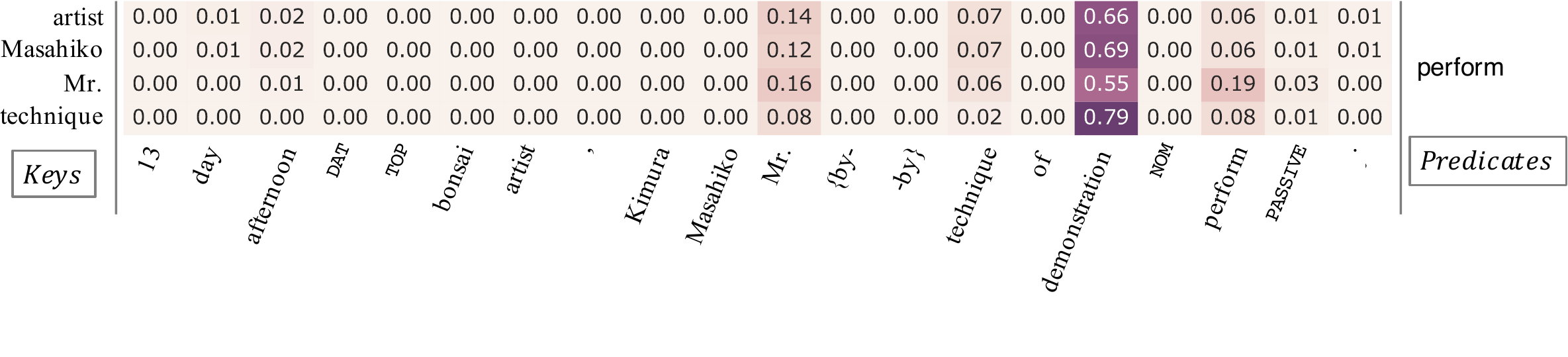}\\
	\caption{Attention weights of {\sc MP-Pool-SelfAtt} for Example~(4).}
	\label{fig:attention-ex4}
\end{figure*}

To analyze the behavior of our proposed models in detail, we show some prediction examples of the {\sc SelfAtt}, {\sc MP-SelfAtt}, and {\sc MP-Pool-Att} models in the development set with the weights in the attention layers in Figures~\ref{fig:analysis}-\ref{fig:attention-ex4}.

In Figure~\ref{fig:analysis}, Examples~(1)~and~(2) are the instances for which only {\sc SelfAtt} failed to predict the answer and for which only {\sc MP-Pool-SelfAtt} correctly predicted the answer, respectively.
For these examples, the weights in the attention layers behave similarly.
Figure~\ref{fig:attention-ex1} shows the weights for Example~(1).
In this sentence, the correct {\tt NOM} of {\it stretch}, {\it Tanigawa}, is also {\tt NOM} of {\it respond}, which is relatively easy to predict.
{\sc SelfAtt}, which is designed to capture dependencies over a single predicate-argument structure, failed to predict {\tt NOM} of {\it stretch} most likely because the answer {\it Tanigawa} is distant from the target predicate with its limited syntactic clues.
Conversely, {\sc MP-Pool-SelfAtt} and {\sc MP-SelfAtt} successfully predicted the answer by taking the answer token {\it Tanigawa} into account when computing the score of the counter candidate {\it thinking}.
{\sc MP-SelfAtt}, the model that incorporates the other predicate positions into {\sc SelfAtt}, significantly increases the weight for the answer token. {\sc MP-Pool-SelfAtt}, which explicitly integrates the predictions for the other predicates, further increases the weight for the answer token.
This example demonstrates that the proposed extensions successfully predict a correct argument by considering the relation to the argument in another predicate where the syntactic relation between the predicate and argument is much clearer and thus the argument relation is relatively easy to predict.
Due to space limitations, we cannot show the weights for Example (2), but the same also holds for that example. {\sc MP-Pool-SelfAtt} focuses on professors, which is the ``easy-to-predict'' NOM argument of dive, when the model computes the scores of this token for {\it take} and, consequently, {\it support}. {\sc SelfAtt} and {\sc MP-SelfAtt} assign smaller weights to that token for {\it take} and even smaller weights for {\it support}, which is far from the answer token.

Examples~(3) and (4) are the instances where all the three models failed to predict the answers.
Figure~\ref{fig:attention-ex3} illustrates the attention weights in {\sc MP-Pool-SelfAtt} for Example~(3). To solve this example, the model is expected to understand that {\tt NOM} of {\it accept} should be the same as the persons who received the order from the {\it ministries}. However, {\sc MP-Pool-SelfAtt} could not acquire this kind of dialog-level knowledge and
pays little attention to the correct argument {\it staff} when the model computes the score of the wrong answer {\it ministries} for {\tt NOM} of {\it accept}.

In Example~(4), {\tt NOM} of the nominal predicate {\it demonstration} can be a clue for predicting {\tt NOM} of {\it perform}. However, the models currently do not predict the arguments of nominal predicates and therefore cannot capture the relationships between these two sufficiently (Figure~\ref{fig:attention-ex4}).
This example suggests one of our future directions: the joint prediction of verbal and nominal predicates.

\section{Related Work}
\paragraph{End-to-End Models in SRL}
End-to-end approaches to SRL have been widely explored recently, and many state-of-the-art results have been achieved \cite{Zhou2015,He2017,Marcheggiani2017,Tan2017}. Following these advanced models, we adopted a stacked bi-RNN as our base model.

\paragraph{Methods for Dealing with Long-Distance Dependencies in End-to-End Models}
In SRL studies, \newcite{Marcheggiani2017} proposed a variant of deep bi-RNN models that connects the intermediate representations of the predictions for the words in syntactic dependency relations on top of the deep RNN.
Very recently, aiming to directly connect the related words, \newcite{Tan2017} stacked self-attention layers, each of which followed a feedforward layer, in a manner similar to the method of \newcite{Vaswani2017}, which was originally applied to an encoder-decoder model.

Self-attention has been successfully applied to several NLP tasks, including textual entailment, sentiment analysis, summarization, machine translation, and language understanding~\cite{Paulus2017,Shen2017,Lin2017,Vaswani2017}.
Techniques using pooling have been applied to merge intermediate expressions in predictions in the tasks where related tokens are often at long distance such as coreference resolution and machine reading~\cite{Clark2016,Kobayashi2016}.
One major contribution of this study is its novel idea of using these techniques for capturing long-distance dependencies for modeling interactions among multiple predicate-argument relations.

\paragraph{Approaches to Capturing Multi-Predicate Interactions}
For Japanese, \newcite{Ouchi2015} jointly identified arguments of multiple predicates by modeling argument interactions with a bipartite graph. \newcite{Iida2015} constructed a {\it subject-shared predicate network} and deterministically propagated the predicted subjects to other predicates. \newcite{Shibata2016a} adapted a NN framework to \newcite{Ouchi2015}'s model using a feedforward network.
For an end-to-end neural model, \newcite{ouchi2017} used a Grid RNN to capture multiple predicate interactions.
Through experiments, we demonstrated that our proposed models outperformed these models in terms of the overall $F_1$ on a standard benchmark corpus.\footnote{\newcite{Shibata2016a} evaluated the model on a different dataset and hence, it is difficult to compare the results directly.}

To the best of our knowledge, there are few previous studies related to SRL considering multiple predicate interactions for languages other than Japanese. \newcite{Yang2014} performed a discriminative reranking in the role classification of shared arguments.
\newcite{Lei2015} proposed an SRL model based on the dimensionality reduction on a tensor representation to capture meaningful interactions between the argument, predicate, corresponding features, and role label.
It is not straightforward to compare these methods with our models; however, it is an intriguing future issue to consider how well the techniques devised for Japanese PAS analysis work for other languages.

\paragraph{Other Approaches to Argument Omission}
In order to perform robust prediction for arguments with fewer syntactic clues,
several previous studies have explored various types of selectional preference scores that consider the semantic relations between a predicate and its arguments~\cite{Iida2007,imamura2009discriminative,komachi2010,sasano2011discriminative,Shibata2016a}.
This direction of research is orthogonal to our approach, suggesting that the models could be further improved by being combined with these extra features.

\section{Conclusion}
In this study, we have proposed new Japanese PAS analysis models that integrate prediction information of arguments in multiple predicates.
We extended the end-to-end style model using a deep bi-RNN and introduced the components that consider the multiple predicate interactions into the input and last layers. As a result, we achieved a new state-of-the-art accuracy on the standard benchmark data.

Our detailed analysis showed that the proposed models successfully predict the correct arguments by using the information of the {\it ``easy-to-predict''} arguments in other predicates. In addition, the error analysis suggests that jointly predicting the arguments of verbal and nominal predicates may further improve the performance.
An intriguing issue we plan to address next is how to extend the proposed interaction layer to cross-sentential interactions of PASs.

\section*{Acknowledgements}
We are grateful to the anonymous reviewers for their useful comments and suggestions.
We thank Hiroki Ouchi for his help in checking our re-implementation.
We also thank Shun Kiyono and Kento Watanabe for valuable discussions.
This work was partially supported by JSPS KAKENHI Grant Numbers 15H01702 and 15K16045.

\bibliographystyle{acl}
\bibliography{acl2018}

\newpage

\section*{Appendix A: Implementation Details}
\label{sec:supplemental}

\paragraph{Hyperparameters}
The hyperparameters were selected to obtain a maximum $F_1$ on the development set.
The dimension of the word embeddings $d_w$ was set to $256$.
The dimension of the hidden state of the GRUs $d_r$ was set to $256$ from $\{128,256,512\}$ and the number of the GRU layers was set to $10$ from $\{6,8,10,12\}$.
The dropout rate of the GRUs was set to $0.1$ from $\{0.0,0.1,0.2\}$.
The dimensions of the outputs of the nonlinear transformations $f$, $g$ and $h_{i,t}$ were set to $1024$ from $\{512,768,1024\}$.
We set the batch size of the training data as the number of predicates in each sentence.
We employed the negative log likelihood as the training loss and an Adam optimizer with $\beta_1=0.9$, $\beta_2=0.999$, and $\epsilon=1e-08$.
During the training, we halved the learning rate when the $F_1$ score on the development set did not improve after four epochs, and restarted training with the parameters that obtained the maximum $F_1$ score.
We repeated this process and terminated the training when the new learning rate was less than 1/16 of the initial value.
The initial learning rate of each model was selected from $\{0.00002, 0.00005, 0.0001, 0.0002, 0.0005\}$. The output threshold for each label $\theta_c\in [0.0, 1.0]$ was searched in increments of $0.01$ to maximize the $ F_1 $ score in the training data.

\paragraph{Preprocessing}
As initial word embeddings, we used vectors obtained via the same procedure as the one proposed by \newcite{Matsubayashi2017} using
Japanese Wikipedia articles. These vectors were fine-tuned in the training.
Following their approach, we used part of speech (PoS) vectors for words that were not contained in the lexicon of the Wikipedia articles. We used the CaboCha parser v0.68~\footnote{\url{https://taku910.github.io/cabocha/}}
with the JUMAN dictionary for word segmentation and PoS tagging of NTC.

\end{document}